# EVO* 2019

# The Leading European Event on Bio-Inspired Computation

*Leipzig, Germany. 24-26 April 2019*

# – LATE BREAKING ABSTRACTS –

Editors:

A.M. Mora
A.I. Esparcia-Alcázar

# Preface

This volume contains the Late- Breaking Abstracts submitted to the EVO* 2019 Conference, that took place in Leipzig, from 24 to 26 of April.

These papers where presented as short talks and also at the poster session of the conference together with other regular submissions.

All of them present ongoing research and preliminary results investigating on the application of different approaches of Evolutionary Computation to different problems, most of them real world ones.

We consider these contributions as very promising, since they outline some of the incoming advances in the area of evolutionary algorithms applications.

<div style="text-align: right;">
Antonio M. Mora  
Anna I. Esparcia-Alcázar
</div>

# Table of contents

# Neural Darwinism as Combinatorial Optimization

Darrell Whitley
Department of Computer Science
Colorado State University
Fort Collins, CO, USA
whitley@cs.colostate.edu

The current success of deep learning has fueled renewed interest in other forms of artificial neural networks. Evolutionary computation in particular and combinatorial optimization in general can play a role in neural network learning and in shaping how neural networks are configured.

Recently, Whitley et al. [1] have converted the neural network learning into a problem of neuron selection. Instead of focusing exclusively on weight optimization, they show that a neural network can learn using randomly generated neurons; the key to this form of learning is to discover useful nonlinear combinations of neurons that can work together.

This idea is closely related to Gerald Edelman's theory of "Neural Darwinism" [2]. Edelman proposes that neuron selection is a key element in how brains are configured and that brain development is strong influenced by a form of group selection that acts on co-adapted neurons. The use of group selection to explain Neural Darwinism is in fact one of the weaknesses of Edelman's theory, since group selection is in fact very rare and unstable. However, Whitley et al. converted the neuron selection problem into a k-bounded pseudo-Boolean optimization problem [3].

But by posing neuron selection as a k-bounded problem, the need for the evolution of neural architectures by group selection goes away. The resulting neural architectures can also be viewed as a spin glass system, where minimizing the potential energy of the system also minimizes the learning error.

We have also proven [4] that every k-bounded pseudo-Boolean optimization problem can be expressed as a discrete Fourier polynomial expressed as the sum of $k$ eigenvectors:

$$f(x) = \bar{f} + \sum_{j=1}^{k} \varphi^{(j)}(x) \qquad (1)$$

where each subfunction is composed of all of the $j^{th}$ order coefficients of the discrete Fourier polynomial and $w_o = \bar{f} = \varphi^{(0)}(x)$.

Each subfunction is also an eigenvector of the Graph Laplacian of the Hamming neighborhood of the search space [5,6].

This opens the door to computing various summary statistics which can characterize the robustness of neural architectures.

This talk will also explore how both past and current research in evolutionary algorithms contributes to current research on artificial neural systems and deep learning.

**REFERENCES**

[1] D. Whitley, R. Tinos, F. Chicano. Optimal Neuron Selection and Generalization: NK Ensemble Neural Networks. In: *International Conference on Parallel Problem Solving from Nature* (PPSN XV). Springer, LNCS volume 11101, pp: 449-460, 2018.
[2] G.M. Edelman, Neural Darwinism: Selection and Reentrant Signaling in Higher Brain Function. *Neuron*, 10(2):115-125, 1993.
[3] E. Boros and P.L. Hammer. Pseudo-Boolean Optimization. *Discrete Applied Mathematics*, 123(1):155–225, 2002.
[4] A.M.Sutton, A.Howe, and D.Whitley. A Theoretical Analysis of the k-satisfiability Search Space. In: *Engineering Stochastic Local Search Algorithms* (LSL 2009), Springer, LNCS, volume 5752, pp: 46-60, 2009.
[5] Lov K. Grover. Local search and the local structure of NP-complete problems. *Operations Research Letters*, 12:235–243, 1992.
[6] Christian M. Reidys and Peter F. Stadler. Combinatorial landscapes. *SIAM Review*, 44:3–54, 2002.

# The Plant Propagation Algorithm on Timetables: First Results

Romi Geleijn
Institute of Interdisciplinary Studies
University of Amsterdam, The Netherlands
romi.geleijn@student.uva.nl

Marrit van der Meer*
Institute of Interdisciplinary Studies
University of Amsterdam, The Netherlands
marrit.vandermeer@student.uva.nl

Quinten van der Post
Minor Programmeren
University of Amsterdam, The Netherlands
q.f.vanderpost@uva.nl

Daan van den Berg
Minor Programmeren, Docentengroep IvI
University of Amsterdam, The Netherlands
d.vandenberg@uva.nl

## ABSTRACT

One Stochastic HillClimber and two implementations of the Plant Propagation Algorithm (PPA-1 and PPA-2) are applied to an instance of the University Course Timetabling Problem from the University of Amsterdam. After completing 10 runs of 200,000 objective function evaluations each, results show that PPA-1 outperforms the HillClimber, but PPA-2 makes the best timetables.

## 1 INTRODUCTION

Universities all over the world are faced with the University Course Timetabling Problem (UCTP), an NP-hard constrained optimization problem, which means that an optimal solution for a realistically sized timetable cannot be found within any reasonable amount of time [1]. For these kinds of problems, exact algorithms are practically useless, but sufficiently good solutions can be produced by heuristic optimization algorithms such as Genetic Algorithms, Ant Colony Optimization and Tabu Search [1] [2] [3] [4] [5] [6].

In this preliminary investigation, performance of the Plant Propagation Algorithm (PPA), a bio-inspired meta-heuristic, is assessed when applied to the UCTP at the University of Amsterdam (UvA). Previous studies have applied PPA to the Uncapacitated Exam Scheduling Problem (UESP), which is related to UCTP as both are a subcategory of Academic Scheduling Problems, but also to the Traveling Salesperson Problem (TSP), another NP-hard constrained optimization problem [7][8]. The algorithm has not been subjected to the UCTP itself, but has performed well on a diverse array of combinatorial optimization problems and might be a promising candidate, given that we develop a suitable adaptation from its previous implementations [9] [10]. This study uses a(n anonymized) dataset from the UvA and evaluates the performance of three algorithms for optimally scheduling its courses. First, a simple Stochastic HillClimber (HC) algorithm (a.k.a. "stochastic local search") is implemented. Second, PPA-1 is a direct adaptation from its TSP-cousin to the timetabling problem [8]. PPA-2 finally, is an adaptation that stems from the seminal implementation of PPA on continuous functions [9], therefrom inheriting a somewhat smoother procreation strategy. The results of all three algorithms on this single real-world instance of UCTP are quantitatively compared.

## 2 TIMETABLES AND OBJECTIVE VALUES

For this explorative study, data from 29 existing courses and 609 fictional UvA-students is used. Every student is enrolled in 1 to 5 courses, and no interdependencies between courses are currently implemented. Courses consist of zero or more plenary lectures, study groups and lab practicals ('labs'), the latter two activities occasionally being split up in equally sized *sessions* to meet capacity constraints. For instance, if there are 78 students enrolled in "Machine Learning II", but labs in this course can only accommodate 20 students at a time, four sessions of that single lab are scheduled to accommodate all enrolled students. All 129 course activities are scheduled in 7 rooms with varying capacities from the UvA's Science Park location, each having 4 time slots on all 5 weekdays, amounting to 140 weekly *room slots*. Neglecting symmetries and equivalences, these numbers of course activities and room slots give rise to $\frac{129!}{11!} \approx 3.4 * 10^{233}$ different timetable configurations just for one week, even for this reduced problem instance.

An initial timetable is created by assigning each course activity to exactly one randomly chosen room slot. Then, every student attending a course is assigned to all activities within the course; if a course activity is split up, one available session is selected at random. Once completed, this constraint-satisfying (or *'valid'*) initial timetable is assigned a base objective value of 1,000, after which three objective modifiers are applied. First, for every activity, any student number that exceeds the room size reduces the objective value by one. Second, for each student that has more than one course activity at any given time slot, one point is deducted for every excess activity. Third, if a course has its activities spread optimally over the week (e.g. two activities either on Monday-Thursday or Tuesday-Friday), 20 points are added to the timetable's objective value. Conversely, if the number of course activities is higher than its scheduled days number, 10 points are deducted for each shortcoming day. If a course has one or more activities split up in sessions which are scheduled on different days, points are attributed relative to the fraction of students for whom the spread is (sub)optimal. From this objective function, every existable valid timetable has an objective value within the upper and lower bounds of 1580 and -6001.

## 3 THREE ALGORITHMS

All three algorithms start off with randomized initial timetables and repeatedly applying swap-mutations, exchanging the contents of two randomly selected room slots, similar to a 2-opt in the TSP.

The HillClimber performs a swap-mutation each iteration, which is reverted only if it lowers the objective value of the new timetable.

PPA-1 keeps a population of 40 individuals in descending order of fitness. Every iteration, each individual produces offspring: 10, 5, 3 and 2 new individuals for its top 10% in the population, all mutated with a single swap-mutation, and 1 new individual for the remaining 90% which is subjected to three successive swap-mutations. From each individual and its *own* offspring (its 'family') the fittest individual remains in the population; the others are discarded. Thereby, PPA-1 is an almost direct translation from PPA for the TSP [8].

For the PPA-2 algorithm, there is no 10% - 90% division, but the number of offspring $n_i$ and their mutability $d_i$ is calculated from the smoother 'normalized fitness' value $N_i$ from an individual's objective value $f(x_i)$ as

$$N_i = \frac{1}{2}(tanh(4 \cdot (\frac{f(x_{max}) - f(x_i)}{f(x_{max}) - f(x_{min})}) - 2) + 1) \quad (1)$$

in which $f(x_{max})$ and $f(x_{min})$ are the highest and lowest objective values in the population. The number of offspring for an individual is $n_i = \lceil n_{max} N_i r \rceil$, which are all mutated as $d_i = \lceil s_{max} \cdot r \cdot (1-N_i) \rceil$ swap-mutations, in which $n_{max}$ denotes the maximum number of swaps per offspring, $s_{max}$ is the maximum allowable number of swap-mutations, and $r$ is a random number in [0,1] which is redrawn every time it is invoked (anywhere). In this experiment, parameters $n_{max} = 10$ and $s_{max} = 20$ were used for all runs of PPA-2. Finally, all the newly generated offspring are added to the population, which is then sorted and retains only the best 40 individuals. Thereby, all three algorithms adopt an *elitist* approach, meaning the best objective value never decreases during a run (figure 1).

## 4 RESULTS, CONCLUSION & DISCUSSION

In this preliminary investigation, PPA-1 finds better timetables than the HillClimber, with maximum, average and minimum objective values of 1250, 1231 and 1209, over 1239, 1223 and 1201 after 10 runs of 200,000 function evaluations. PPA-2 performs best (1263, 1246, 1235), surprisingly outperforming both other algorithms after as many as 100,000 function evaluations. The absolute differences however are small and the rapid convergence of the HillClimber, traditionally susceptible to local maxima, could indicate that the objective function from this problem instance has a high degree of convexity. Contrarily, the fact that both PPA-algorithms – with their high-mutability offspring – persistently surpass the HillClimber after a very long time might indicate that local maxima are few and far between, which in terms might be due to the sheer vastness of this problem's state space [11].

In short, both PPA-algorithms seem to be plausible candidates for NP-hard optimization problem instances other than the TSP, such as this instance of the UCTP. Nonetheless, the gargantuan computational effort required for a better-than-HillClimbing solution raises some serious questions about the optimal parameterization of the proposed implementations. Furthermore, a more detailed state space survey including convexity measures, saddle-node maxima detection or symmetry-breaking might seriously improve final solutions, speed of convergence and total runtime.

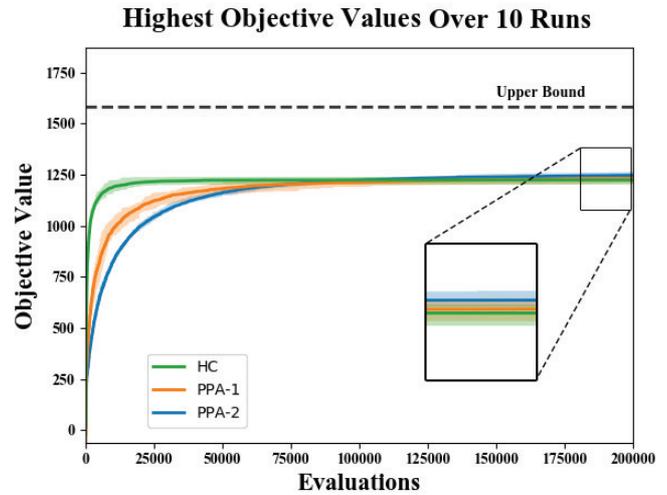

**Figure 1:** Though the HillClimber rapidly finds good solutions, PPA-1 and PPA-2 produce better timetables, but only after about 100,000 function evaluations. Solid lines are averages, transparent areas show the min/max objective values.

## REFERENCES

[1] Pupong Pongcharoen, Weena Promtet, Pisal Yenradee, and Christian Hicks. Stochastic optimisation timetabling tool for university course scheduling. *International Journal of Production Economics*, 112(2):903–918, 2008.

[2] Hamed Babaei, Jaber Karimpour, and Amin Hadidi. A survey of approaches for university course timetabling problem. *Computers & Industrial Engineering*, 86:43–59, 2015.

[3] Edmund K Burke, Graham Kendall, Mustafa Mısır, and Ender Özcan. Monte carlo hyper-heuristics for examination timetabling. *Annals of Operations Research*, 196(1):73–90, 2012.

[4] Olivia Rossi-Doria, Michael Sampels, Mauro Birattari, Marco Chiarandini, Marco Dorigo, Luca M Gambardella, Joshua Knowles, Max Manfrin, Monaldo Mastrolilli, Ben Paechter, et al. A comparison of the performance of different metaheuristics on the timetabling problem. In *International Conference on the Practice and Theory of Automated Timetabling*, pages 329–351. Springer, 2002.

[5] Krzysztof Socha, Michael Sampels, and Max Manfrin. Ant algorithms for the university course timetabling problem with regard to the state-of-the-art. In *Workshops on Applications of Evolutionary Computation*, pages 334–345. Springer, 2003.

[6] Chong Keat Teoh, Antoni Wibowo, and Mohd Salihin Ngadiman. Review of state of the art for metaheuristic techniques in academic scheduling problems. *Artificial Intelligence Review*, 44(1):1–21, 2015.

[7] Meryem Cheraitia, Salim Haddadi, and Abdellah Salhi. Hybridizing plant propagation and local search for uncapacitated exam scheduling problems. *International Journal of of Services and Operations Management.*, 2017.

[8] Birsen İ Selamoğlu and Abdellah Salhi. The plant propagation algorithm for discrete optimisation: The case of the travelling salesman problem. In *Nature-inspired computation in engineering*, pages 43–61. Springer, 2016.

[9] Abdellah Salhi and Eric S Fraga. Nature-inspired optimisation approaches and the new plant propagation algorithm. 2011.

[10] Muhammad Sulaiman, Abdellah Salhi, Birsen Irem Selamoglu, and Omar Bahaaldin Kirikchi. A plant propagation algorithm for constrained engineering optimisation problems. *Mathematical Problems in Engineering*, 2014, 2014.

[11] Misha Paauw and Daan Van den Berg. Paintings, polygons and plant propagation. Springer, 2019.

# In the Quest for Energy Efficient Genetic Algorithms

F. Fernández de Vega, J. Díaz, J.A. García, F. Chávez, J. Alvarado [†]
Centro Universitario de Mérida
University of Extremadura
Merida, Spain
{fcofdez, fchavez, mjdiaz, jangelgm}@unex.es

**ABSTRACT**

Although usually quality of solutions and running time are the main features of algorithms, recently a new trend in computer science tries to contextualize these features under a new perspective: power consumption. This paper presents a preliminary analysis of the standard genetic algorithm, using two well-known benchmark problems, considering power consumption when battery-powered devices are used to run them. Results show that some of the main parameters of the algorithm has an impact on instantaneous energy consumption -that departs from the expected behavior, and therefore on the amount of energy required to run the algorithm. Although we are still far from finding the way to design energy-efficient EAs, we think the results open up a new perspective that will allow us to achieve this goal in the future.

**KEYWORDS**

Energy Efficient Algorithms; Genetic Algorithms.

## 1. CONSIDERING POWER CONSUMPTION IN GAS

Since the beginning of GAs, this evolutionary based search and optimization heuristic has been run by researchers in any available hardware device that allows them to obtain solutions of quality as soon as possible. However, the algorithm power consumption has never been considered as something of interest, although in other computer science areas the topic has already enter the optimization arena [1], [2], [3].

During the last couple of years, the first attempts to study energy consumption related behaviors in evolutionary algorithms have already been published [4], that includes some preliminary analysis of power consumption associated to different hardware platforms [5]. This study is particularly relevant when battery powered devices (such as hand-held ones or laptop computers disconnected from the mains) are used, for obvious reasons. Yet, to the best of our knowledge no specific study has been presented that analyzes the impact of the algorithm configuration on the energy consumed to reach a solution.

This paper thus presents for the first time such an analysis for GAs. Although results are still preliminary, we consider they pave the way to a better understanding of the algorithm under this new perspective, that will allow in the future the design of more energy efficient GAs.

### 1.1. Population Size, Chromosome size and energy consumption.

Number of generations, population size and fitness functions are the key components that influences the run time of a standard Genetic Algorithm.

If we thus decide to run the algorithm for a previously established number N of generations, it will take shorter than N+1, and longer than N-1. We do not consider here that the solution is found before that number of generations, which can be easily assured by making the problem hard enough (for instance, increasing chromosome size in the *maxone* problem). Something similar can be said regarding the number of individuals in the population. Taking this into account, let us consider that the CPU, when running the algorithm, devotes exactly the same effort regardless the specific operation it is performing. This means that the instantaneous energy consumption is the same along the experiment. Although no previous study on this issue has been performed for EAs, given that no interest on the topic has been described yet on the literature, we can safely state that everybody has assumed that there is no actual difference on the way CPUs work on a given experiment along all the time it is run, regardless of the specific operation performed at any given moment. The EA community is thus implicitly assuming that the important measure is the total power consumption along a given experiment, and it could easily be computed multiplying running time and instantaneous energy consumption of the processor at any given time. We thus will focus here on total power consumed by the GA.

Assuming the previously described considerations, we could easily build a graph showing power consumption estimated for different parameter values in the GA. For instance, if a population with size N (N individuals) consumes a given amount of energy during a run, if we expand the population size in a series of experiments, we expect to have power consumption values proportional to the expansion, given that the algorithm will perform as many extra operations (fitness evaluations, mutations, crossover...) proportional to the new number of individuals in the population. We show the kind of expected behavior in figure 1.

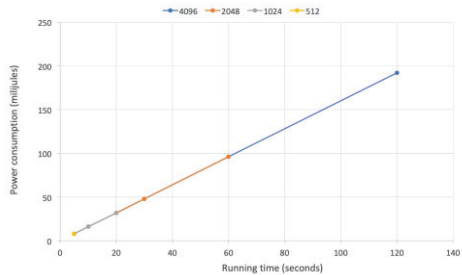

**Figure 1: Expected GA power consumption with different population sizes.**

Result has been obtained by using values for a single experiment and projecting them to the remaining ones, given that the same behavior should be expected regardless of the population size. We have also considered that solution is not found along the run. The main idea used to build such a graph is that processor's instantaneous energy consumption is constant regardless of the operation performed: the longer the time to run a given experiment, the higher the energy required to run it (energy and time proportional values). Of course, better fitness values will be found with larger populations, but that is not the point considered here. We hypothesize that this expected behavior is what has hinder researchers from taking it into account when studying the behavior of the Genetic Algorithm: if running time and power consumption are proportional values, there is not reason to study both. Once running time is obtained, power consumption can be easily derived. Although that maybe the case, it has recently been described that energy is important to decide the more efficient hardware platform to run an algorithm [4]. I any case, should we still assume energy and time are proportional values of a single entity without experimental evidence? Do some of the GA main parameters somehow influence energy required for the algorithm to run? We try to answer these question in the next section with a preliminary analysis that addresses these issues.

## 2. ANALYSIS OF GAS POWER CONSUMPTION.

We have performed such an analysis using wo well known GA benchmark problems: maxone and trap functions, with different chromosome and population sizes (as seen in figures that follows). The idea is to have different configurations, using some of the main GA parameters when long runs of the experiments are performed, and then compute total energy consumed when different parameters settings are applied. Generational version of the algorithm is run, with a maximum time limit established for the run: 300 seconds; population size and problem difficulty (chromosome size) were checked for the maxone problem, and then, the trap function tested to confirm some of the conclusions draw. For each of the parameter values 30 independent runs were launched so that averaged values are shown below. Despite the difficulty established for the problems, some of the runs using large populations were able to find the solution before reaching 300 seconds, so in that cases mean values were computed for the runs that reached that time steps. In this first study we have chosen a Lenovo Tablet Tab2, A10 - 70F with a Mediatek SoC MT8165, which is a quite standard representative model among the options available.

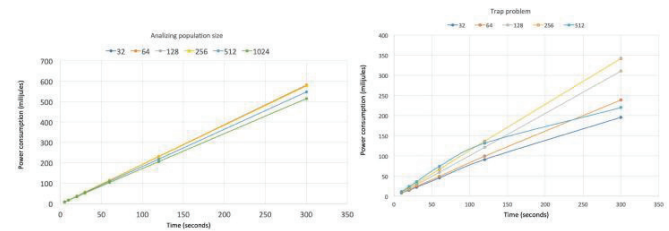

**Figure 2: GA power consumption with different population sizes.**

Figure 2 clearly show that some "anomalies" have been detected in these series of experiments (when compared to expected results in figure 1). We can confirm that for the first time we have detected that some of the main parameter values, has an effect on the power consumption of the algorithm; particularly relevant is population size employed. This could be related to memory usage patterns, that may affect cache access operations.

## 3. CONCLUSIONS

Energy consumption patterns have been analyzed for the maxone problem and the trap functions. Results show that non-linear behaviors can be found when we analyze relationships among chromosome sizes or number of individuals in the population and the energy required to complete the experiment. These anomalies -such as smaller consumption with larger population sizes- deserves further research in the future.

### ACKNOWLEDGMENTS
We acknowledge support from Spanish Ministry of Economy and Competitiveness under project TIN2017-85727-C4-f2,4g-P, Regional Government of Extremadura, Department of Commerce and Economy, the European Regional Development Fund, a way to build Europe, under the project IB16035 and Junta de Extremadura, project GR15068.

### REFERENCES
[1] Ye, M. et al. "EECS: an energy efficient clustering scheme in wireless sensor networks." Performance, Computing, and Communications Con-ference, 2005. IPCCC 2005. 24th IEEE International. IEEE, 2005.
[2] Camilo, T., Carreto, C., Silva, J. S., & Boavida, F. (2006, September). An energy-efficient ant-based routing algorithm for wireless sensor networks. In International Workshop on Ant Colony Optimization and Swarm Intelligence (pp. 49-59). Springer, Berlin, Heidelberg.
[3] Albers, S., Energy-efficient algorithms. Communications of the ACM, 53(5):8696, 2010.
[4] Fernández de Vega, F., Chávez, F., Díaz, J., García, J. A., Castillo, P. A., Merelo, J. J., & Cotta, C. (2016, September). A cross-platform assessment of energy consumption in evolutionary algorithms. In International Conference on Parallel Problem Solving from Nature (pp. 548-557). Springer, Cham.
[5] Alvarez, J. D., Chavez,´ F., Castillo, P. A., Garc´ıa, J. A., Rodr´ıguez, F. J., & Fernandez´ de Vega, F. (2018). A Fuzzy Rule-Based System to Predict Energy Consumption of Genetic Programming Algorithms. Computer Science & Information Systems, 15(3).

# Deep evolutionary training of a videogame designer

Álvaro Gutiérrez-Rodríguez
ETSI Informática
Málaga, Spain
alvarogr@lcc.uma.es

Carlos Cotta
ETSI Informática
Málaga, Spain
ccottap@lcc.uma.es

Antonio J. Fernández-Leiva
ETSI Informática
Málaga, Spain
afdez@lcc.uma.es

## ABSTRACT
This work presents a procedural content generation system that focuses on the design of levels in Metroidvania games using a model of the preferences and experience of the designers. This model is subsequently exploited by an optimization component that tries to create adequate game designs. By iterating over this process, the model is augmented with Artificial Intelligence (AI)-generated data. We focus on the influence in the system output of factors such as the composition of the initial training set and the potential intervention of the user during the process. The experimental results show how the diversity of the former is essential for performance, and how the participation of the human expert can result in more focused designs.

## KEYWORDS
Procedural content generation, Neural networks, Genetic algorithm, Game design

## 1 INTRODUCTION

One of the problems in videogames is the monotony in the gameplay, this makes players lose interest in replaying the game once they complete all the levels in it. On the other hand, the work carried out by design teams in videogame studios is very arduous, expensive with deadlines to finish their work, adding a stress component. This greatly reduces the concentration and contribution of new ideas, which is reflected in the quality of the designs.

For alleviating the monotony problem, and for assisting in the development phase, procedural content generation (PCG) can constitute an essential tool. Real examples of this generation can be found from procedural generation of maps[1] or weapons[2] to stories[3].

The field of intelligent assistance tools for the design of videogames [1] have been explored with promising results [2, 2–4]. Given the difficulty of the problem, most of these works have focused on the generation of levels, for a specific genre and in stablished mechanics. In this article, we extend a previous work where we proposed an Artificial Intelligence (AI)-assisted design tool for videogames which can generates a complete games for the videogame genre Metroidvania [5]. Our proposal uses evolutionary algorithms (EAs) for the generation of content and employs artificial neural networks (ANN) to mimic the way the designer thinks.

We deploy here the system on a realistic environment where the problems of lack of initial information and imbalance of learning cases are found. In the scientific literature, these problems are addressed using data augmentation techniques which generate synthetic cases from an initial learning set. Several techniques with promising results have been developed in the generation of images [6]. In this paper, we extend our previous work with a particular data augmentation approach for the generation of synthetic cases of video game levels created by a designer.

## 2 SYNTHETIC GENERATION PROCESS

This process –shown in Fig. 1– trains the classifier with an initial subset of the global reference solutions, and uses the learned model inside the optimization component, both for maximization ($EA^+$ – generation of good solutions) and minimization ($EA^-$ – generation of bad solutions). These solutions are added to the training set with the label of its provenance and the ANN is re-trained with this updated training set. This constitutes an iteration of the system until a maximum of $M$ iterations. This loop is enhanced by considering the potential intervention of the designer in order to filter out the new solutions. At the very end of the process, the final model learned is again used by an EA (analogous to $EA^+$) in order to construct the final designs.

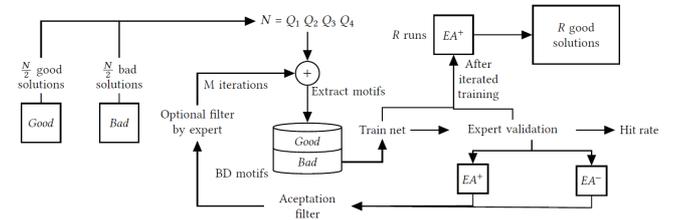

Figure 1: Synthetic generation process scheme

The intervention of the designer (optional filter by expert, in Fig. 1) consists on conduct a new labelling of the generated solutions to prevent erroneous labels by the model.

## 3 EXPERIMENTAL RESULTS

We consider motifs of size $k = 3$, $B = 7$ different types of building blocks and the size of the main path of the level $m = 50$, resulting in a solution space of size $|B|^m \geqslant 10^{42}$. The pattern to learn is constructed by randomly assigning weights between 1 and −1 to each motif. We consider an ANN with an input layer of $k|B| = 21$ neurons, one hidden layer of $n_h = 63$ neurons (3×input layer) and an output layer of $n_o = 2$ neurons (all neurons with sigmoid activation and $\alpha = 0.33$). The ANN is trained using backpropagation with learning rate of $\delta = 0.2$, $momentum = 0$, batch size of 25 and 500 learning epochs.

We executed two EAs: one to generate synthetic learning cases (in-loop), and another to generate solutions at the end of the process (post-loop). The configuration of both EAs is the same, individuals is $\mu = 50$, with a crossover probability of $p_c = 0.9$, mutation of $p_m = 1/m = 0.02$, but they differ in the number of evaluations: *maxevals*= 1000 for in-loop EAs and *maxevals*= 5000 for the post-loop EA.

---
[1] http://spelunkyworld.com/index.html
[2] https://borderlandsthegame.com/
[3] https://www.shadowofwar.com/es/

Table 1: Percentage of solution space correctly evaluated. We report the median ($\widetilde{x}$), mean ($\bar{x}$), and standard error of the mean ($\sigma_{\bar{x}}$) for the 30 executions of each experiments. Symbols •/∘ indicate statistical significance/no statistical significance/ ($\alpha = 0.05$), ★ best overall results and – comparisons of a table entry with itself

|       | 0 interventions | | | 5 interventions | | | 10 interventions | | |
|-------|-----------------|---|---|-----------------|---|---|------------------|---|---|
|       | $\widetilde{x}$ | $\bar{x} \pm \sigma_{\bar{x}}$ | | $\widetilde{x}$ | $\bar{x} \pm \sigma_{\bar{x}}$ | | $\widetilde{x}$ | $\bar{x} \pm \sigma_{\bar{x}}$ | |
| $Q_1$ | 244   | $70.93 \pm 1.12$ | – • • | 250.5 | $72.40 \pm 0.74$ | • • • | 251 | $73.18 \pm 0.00$ | • • • |
| $Q_2$ | 270   | $78.59 \pm 0.69$ | – • • | 271   | $78.65 \pm 0.36$ | ∘ • • | 271 | $79.01 \pm 0.00$ | • • • |
| $Q_3$ | 284   | $82.72 \pm 0.61$ | – • • | 284   | $82.79 \pm 0.35$ | ∘ • • | 284 | $82.79 \pm 0.03$ | ∘ • • |
| $Q_4$ | 291   | $84.48 \pm 1.03$ | – – • | 292.5 | $85.17 \pm 0.58$ | ∘ – • | 295 | $86.01 \pm 0.00$ | • – ★ |

We selected 10 iterations with 3 different values of interventions: 0, 5 and 10. We established a solution acceptance range between 50% and 60% percent of the maximum fitness (or negative maximum, for the EA$^-$).

Each time the model is trained, we keep track of the number of motifs whose evaluation has the same sign as indicated by the underlying model (to be discovered) and the model learned. This evaluation is normalized by the total number of motifs ($|B|^k$), obtaining the percentage of them space correctly evaluated. We established 20 iterations for the post-loop EA.

Twelve different experiments have been executed and each of them has been executed 30 times to study in more detail the behavior trend of the process. The objective is to assess the impact of the size of the initial reference set and the number of designer interventions during the process. The global reference

The global set of reference solutions has 1920 motifs (960 *good* and 960 *bad*) and we sorted in ascending order and partitioned them into 4 quartiles: $Q_1$, $Q_2$, $Q_3$, and $Q_4$, each of them being the merger of the previous quartiles with different sizes of initial training set.

Table 1 shows the percentage of the motif space correctly assessed by the classifier at the end of the synthetic generation process. In general, we can see that the process generates a valid model as measured as the number of motifs evaluated correctly. The intervention of the designer helps to consolidate the fit of the model. The standard error of the mean shows this behavior and the standard error is consistently being reduced with different number of user interventions. If we look at the results by fixing the number of user interventions and varying the training set, the average improvement is more remarkable.

Most of the results are statistically significant. This confirms the capability of the system to generate varied solutions. Some exceptions are found for the intermediate number of user interventions, which does not seem to improve the results unless the initial training set is very restrictive itself. We hypothesize that larger diverse initial set of experiences may be richer and hence require more intense intervention.

## 4 CONCLUSIONS

The results show that if the database of experiences used as starting point is large and diverse enough, then acceptable results can be obtained even in absence of feedback from the designer. For future work, we consider extending the feedback procedures available to the designer so that in addition to filtering out motifs, they can also produce new information dynamically inspired by the system outcome.

## ACKNOWLEDGMENTS

This work is supported by Spanish Ministerio de Economía, Industria y Competitividad under project DeepBIO (TIN2017-85727-C4-1-P), and by Universidad de Málaga, Campus de Excelencia Internacional Andalucía Tech.

## REFERENCES

[1] Georgios N. Yannakakis and Julian Togelius. A panorama of artificial and computational intelligence in games. *IEEE Transactions on Computational Intelligence and AI in Games*, 7(4):317–335, 2015.
[2] Nathan Sorenson and Philippe Pasquier. The evolution of fun: Automatic level design through challenge modeling. In Dan Ventura, Alison Pease, Rafael Pérez y Pérez, Graeme Ritchie, and Tony Veale, editors, *International Conference on Computational Creativity*, pages 258–267, Lisbon, Portugal, 2010. Department of Informatics Engineering, University of Coimbra.
[3] Antonios Liapis, Georgios N. Yannakakis, and Julian Togelius. Sentient sketchbook: Computer-aided game level authoring. In Georgios N. Yannakakis, Espen Aarseth, Kristine Jørgensen, and James C. Lester, editors, *ACM Conference on Foundations of Digital Games*, pages 213–220, Chania, Crete, 2013. Society for the Advancement of the Science of Digital Games.
[4] Michael Cook and Simon Colton. Ludus ex machina: Building a 3d game designer that competes alongside humans. In Simon Colton, Dan Ventura, Nada Lavrač, and Michael Cook, editors, *Fifth International Conference on Computational Creativity*, pages 54–62, Ljubljana, Slovenia, 2014. computationalcreativity.net.
[5] A. Gutiérrez, C. Cotta, and A.J. Fernández-Leiva. An evolutionary approach to metroidvania game design. In *2018 Spanish Conference on Artificial Intelligence*, pages 518–523, Granada, Spain, 2018. Springer.
[6] Joao Correia, Penousal Machado, Juan Romero, and Adrian Carballal. Evolving figurative images using expression-based evolutionary art. In Mary Lou Maher, Tony Veale, Rob Saunders, and Oliver Bown, editors, *Fourth International Conference on Computational Creativity*, pages 24–31. computationalcreativity.net, 2013.

# Smart Chef: Evolving Recipes

Carsten Draschner, Jens Lehmann, Hajira Jabeen
{draschne,jabeen,jens.lehmann}@cs.uni-bonn.de
University of Bonn, Germany

## ABSTRACT
Smart Chef demonstrates the creativity of evolution in culinary arts by autonomously evolving novel and human readable recipes. The evolutionary algorithm for Smart Chef fully automatized and does not require human feedback. The tree representation of recipes is inspired by genetic programming and is enriched with semantic annotations extracted from known recipes. The fitness identifies valid recipes and novelty. Recipe mutation exchanges ingredients by food category classification and recombination interchanges partial recipe instructions. Smart Chef has been tested on a population size of 128 and evolved for 100 generations resulting in valid and novel recipes.

## KEYWORDS
evolutionary algorithm, artificial creativity, recipe, culinary, semantic creativity, genetic programming, food graph, recipe annotation, human readable recipe representation

## 1 INTRODUCTION

Computational creativity is an emerging branch of artificial intelligence that places computers in the center of the creative process. The recently published approaches are focused on selected domains like Graphics or Music generation. Food is an essential part of our life and the dishes we eat are created using various recipes. These recipes demonstrate creativity in combination of ingredients, methods, tastes, textures and proportions. Smart Chef presents an automated system capable of creating novel human readable recipes using Evolutionary Algorithm(EA) for recombining recipes from different regional cuisines.

## 2 RECIPE AS TREES - MACHINE READABLE DATA REPRESENTATION

The initial recipes are fetched from theMealDB.com [1] JSON API. The phenotype to genotype mapping is semi automated. It creates a semantic annotated tree structure inspired by genetic programming (see Figure 1, 2.A, 2.B). The root node represents the final recipe. Each inner node is a granular task (grey) manually constructed based on the preparation sentences. These nodes are assigned a *node-type* and *instruction-type* and attributes, if they are mentioned in the sentence (see Fig. 1). The child nodes of instruction-nodes are either inner instruction nodes (grey) or leaves (blue) identified in the sentence. The leaves are ingredients which have a *name* and a *proportion* and are automatically generated from the recipes ingredient table (see figure 1).

## 3 RECIPE GENERATION USING EVOLUTIONARY ALGORITHM

The evolutionary algorithm generates new recipes in each generation. All the steps of the EA are designed to be fully autonomous for an arbitrary population size.

**INITIALIZATION:** The initial population consists of 128 pre-processed annotated recipe-instruction-trees(see Section 2 and Fig. 1) from theMealDB.com[1]. This population is further evolved to discover novel recipes.

**FITNESS EVALUATION:** The fitness of each recipe is calculated automatically. For this purpose, multiple characteristics of the recipe are taken into account and compared to known recipes. These characteristics are extracted from known valid theMealDB recipes. The recipe normality regarding number of ingredients and procedure steps (related to effort) from the novel recipe are compared to known recipes. The ingredient composition (main-, side-ingredients, spices) patterns are compared the common pattern from known recipes. The children recipes might use same ingredients multiple times in same recipe which is punished in the fitness function. The ingredient-set similarity compared with known recipes is used to give new ingredient combinations a higher creativity score. All those criteria extracted from recipes go into a weighted sum which defines fitness value of each recipe.

**SELECTION** We have used tournament selection based on the assigned fitness value for fitness evaluation.

**RECOMBINATION:** The crossover combines two recipe trees. From the one parent recipe tree a random subtree is replaced by a subtree from the second recipe with similar characteristics (same sub-tree-size). These novel recipes create the next generation of EA for evolution.

**MUTATION:** In each child recipe, one ingredient is replaced by an ingredient from food-databases[2, 3]. The chance for each ingredient in this database to be used is dependent on the food similarity (see Fig.2.D) based on its hierarchical classification (i.e. in foodsubs[3] e.g. Spaghetti has hierarchical classification: Pasta-Rods, Pasta, Grain-Products, Food).

**RESULTING POPULATION** The EA runs for a certain number of generations. For the final population a fully automated genotype - phenotype mapping creates a human readable recipe descriptions with a recipe title, an ingredient table with proportions and fulltext-instructions in natural language (see Figure 1)

## 4 CONCLUSION

This work shows that novel recipes can be fully automatically generated from a genetic programming inspired approach. The proposed genotype-phenotype mapping creates common recipe structures from the recipe trees. For future work, this concept allows arbitrary extensions for fitness evaluation (e.g. recipe: price, sustainability, diet/heath etc.).

Evostar 2019, 24–26 April 2019, Leipzig, Germany. Carsten Draschner, Jens Lehmann, Hajira Jabeen

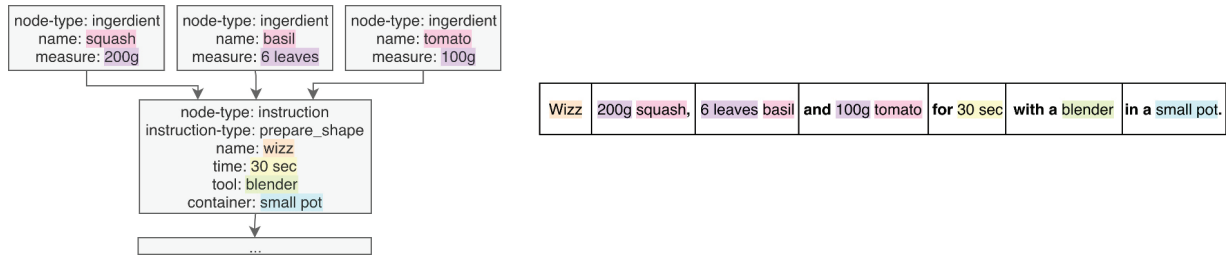

**Figure 1: Example sentence for instruction representations of a recipe. annotated tree structure (left), human readable sentence (right). The phenotype genotype mapping is semi-automatized. The genotype phenotype mapping is fully automatized.**

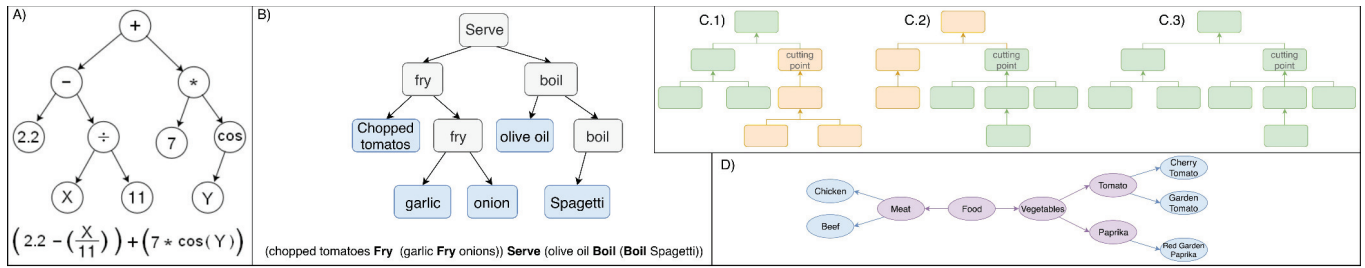

**Figure 2: Genetic Programming Example (A), Corresponding simplified Recipe Example (B), Recombination of Trees (C), Ingredient Hierarchy Example (D)**

## REFERENCES

[1] Alias: zag@kodi.tv. themealdb. https://www.themealdb.com/, 2016-2019. Accessed: 2019-03-10.
[2] The Metabolomics Innovation Centre (TMIC). foodb.ca. foodb.ca, 2017. Accessed: 2018-01-21.
[3] Lori Alden. Foodsubs - the cook's thesaurus. http://www.foodsubs.com/, 1996-2006. Accessed: 2018-01-21.

# Towards Improved Evolutionary Learning of Probabilistic Context-Free Grammars for Protein Sequences

Robert Kowalski
Politechnika Wrocławska
229715@student.pwr.edu.pl

Mateusz Pyzik
Politechnika Wrocławska
mateusz.pyzik@outlook.com

Witold Dyrka
Politechnika Wrocławska
witold.dyrka@pwr.edu.pl

## ABSTRACT
Learning grammatical models is a significant application of evolutionary algorithms. Modeling complex linguistic structures, such as syntax of natural languages and biopolymers requires grammars beyond regular. In the latter field, arguably most successful to date are probabilistic context-free grammars (PCFG), which have been applied to, e.g. RNA structure prediction and - to much less extent - protein sequence analysis. In the most simple case, learning PCFG is confined to estimating probabilities for a fixed set of rules from a positive sample. This is most often achieved using the Inside-Outside algorithm [1]. However, since the procedure is not guaranteed to find the optimum solution [2], alternative heuristic methods gained considerable interest, including genetic algorithms (GA). While some of them allow learning rules together with their probabilities [3], other assume a fixed covering set of rules [4, 5]. Here, we propose a new variant of GA for training PCFG based on a covering set of rules, and compare it with our previous approach [5, 6]. Evaluation is conducted using two toy languages beyond regular, and a bioinformatic set of amino acid sequences.

## KEYWORDS
genetic algorithm, grammar inference, probabilistic context-free grammar, protein sequence

## 1 METHODS

Context-free grammar (CFG) is a quadruple $G = \langle \Sigma, V, v_0, R \rangle$, where $\Sigma$ (alphabet), is a set of terminal symbols, $V$ is a set of non-terminal symbols (variables) disjoint from $\Sigma$, $v_0 \in V$ is a start symbol, and $R$ is a set of production rules rewriting variables into strings of variables and/or terminals. Probabilistic context-free grammar $\mathcal{G} = \langle G, \theta \rangle$ assigns each production rule $r_i \in R$ a corresponding probability $\theta_i \in \theta$. If rule probabilities sum up to 1 over rules rewriting the same variable, the probabilistic grammar is called *proper*. A complete derivation is a chain of rules beginning with $v_0$ and finishing with a string of terminal symbols called a sentence (or sequence in the bioinformatic context). The probability of derivation is the product of probabilities of rules involved. In turn, probability of a sentence $x$ given $\mathcal{G}$ is the sum over all derivations that generate $x$. The grammar is called *consistent* if the probability mass distributed by the grammar over all sequences sums up to 1.

In this piece of research we only consider CFG in the Chomsky Normal Form (CNF), which implies that production rules are either in the form $A \rightarrow a$ (lexical rules) or $B \rightarrow CD$ (structural rules), where lowercase letters denote terminal symbols while uppercase - non-terminal symbols. In addition, we require that a set of variables rewritten with the lexical and structural rules are disjoint (the variables are called lexical and structural, respectively). We call such a grammar form *bipartite* CNF.

**Protein Grammar Evolution** (PGE) is a recently updated framework for evolutionary learning the probabilities of PCFG specialized for amino acid sequences [5, 6] written in C++ using GAlib, Eigen and OpenMP. **Machine Learning for Grammatical Descriptors** (MLGD) is a new alternative scheme [7] in Matlab.

The two frameworks use the Pittsburgh approach, where each individual represents a whole grammar. Technically, an individual $z$ is a real number vector representing probabilities of rules in a given covering set. The individuals are initialized randomly from the interval $[0, 1]$. Before the evaluation step, gene values are normalized to obtain the set of probabilities $\theta(z)$ assuring that the grammar is proper. Given a training set $X$ and an underlying non-probabilistic grammar $G$, the objective function is defined as $f(z) = \frac{1}{|X|} \log \prod_{x \in X} \text{prob}\,[x|\langle G, \theta(z) \rangle]$. In addition, PGE scales the objective function using the triangular sharing method to promote speciation [8], yet its impact is adaptively decreased with the increasing convergence [5]. PGE and MLGD differ also in the selection method which is the tournament out of two for the former, and the roulette wheel for the latter. Both implementations use the steady-state 50% replacement strategy. The two frameworks diverge most significantly in terms of genetic operators. While PGE uses the blended average [8], MLGD applies the one-point crossing-over. Moreover, while the mutation operator of PGE simply draws a new gene value from the uniform distribution $U(0, 1)$, MLGD increases or decreases the gene with a value from $U(0.01, M)$, where $M$ is a scaling parameter. Notably, in MLGD the gene value is set to zero if negative, which effectively conceals the corresponding rule. Finally, only PGE implements convergence as the stop criterion.

## 2 TEST DATA

The learning frameworks were first evaluated using two toy languages over alphabet $\Sigma = \{a, b, c\}$ proposed in [9]. Language $L_1$ is defined as $\{a^n b^n c^m\}$, $n, m \geq 1$. It can be described with a non-probabilistic CFG in bipartite CNF consisting of rules $\{S \rightarrow TC|SC, T \rightarrow AV|AB, V \rightarrow TB, A \rightarrow a, B \rightarrow b, C \rightarrow c\}$, where $S$ is the start symbol. The test set for the non-probabilistic grammar [10] was used as the training set: {*abc*, *aabbc*, *aaabbbc*, *abcc*, *abccc*, *aabbcc*, *aaabbbcc*, *aaaabbbbc*}. The training set, and the positive and negative test sets (76 and 100 sentences, respectively) were generated with http://lukasz.culer.staff.iiar.pwr.edu.pl/gencreator.php.

Language $L_2$ is defined as $\{ac^m\} \cup \{bc^m\}$, $m \geq 1$, and a corresponding minimal CNF CFG is made of rules $\{S \rightarrow AB|SB, A \rightarrow a|b, B \rightarrow c\}$. The training set {*ac*, *bc*, *acc*, *bcc*, *accc*}, and the positive and negative test sets (10 and 100 sequences, respectively) were made with the same tool.

Next, the frameworks were assessed with reference to a set of amino acid sequences of Calcium and Manganese (CaMn) binding site motifs from legume lectins from [6]. The training sample included 24 sequences with at most ca. 70% mutual identity, each 27 amino-acids long. The negative test set consisted of random 100 fragments from the negative test set from [6]. The experiments were carried out in the 4-fold cross-validation scheme.

## 3 EARLY RESULTS

The $L_1$ language was chosen to test parameters of GA including the population size $p = \{40, 80, 160\}$, the crossing-over probability $c = \{0.1, 0.5, 0.9\}$, the mutation probability $m = \{0.01, 0.001, 0.0001\}$ and scale $M = \{0.1, 0.5, 0.9\}$ (only MLGD); middle values were used as the baseline for all population sizes. Each setup was run three times. The covering set of rules included all combinations of 3 lexical and 4 structural variables possible in the bipartite CNF (9 lexical and 196 structural rules). The GA was expected to find a PCFG with the objective log score close to that achievable with the concise grammar used to generate the training set (perplexity per letter of 0.43 nats). Indeed, MLGD was able to find such solutions in 63% cases spread over almost all setups. This is in contrast to PGE, which achieved the goal in just 6% cases, requiring larger population and high crossing-over rate. Five best-fitting grammars in each framework were small (12-13 rules for MLGD and 11-16 rules with probability above 0.01 for PGE). When used as classifiers, all grammars except few with $m = 0.0001$ achieved AuROC above 0.95, with MLGD typically requiring less epochs.

The $L_2$ language was used to test if the frameworks can effectively drop potentially redundant non-terminal symbols. The parameters of GA included the population size of 40 and the middle values of the other parameters (except $c = 0.9$ for PGE). Four covering sets included all combinations of bipartite CNF rules for $2 + 1$, $3 + 1$, $2 + 2$, and $2 + 3$ lexical and structural variables, resp. Each setup was run three times. Again, MLGD achieved higher objective log scores and lower numbers of rules (with probability above 0.01) than PGE. For MLGD, the size of grammars was 5, 8 and ca. 20 rules for 3, 4 and 5 variables, resp. Notably, using additional structural variables improved the objective log score.

Eventually, the CaMn language was used to assess how these results translate to performance in a bioinformatic setting. The parameters of GA were the same as for $L_2$. MLGD was run once for each fold, while PGE was run 56 times accordingly to a slightly more complex partitioning of the sample [6]. Consistently with the previous tests, MLGD outperformed PGE in terms of the objective log score (perplexity of 2.5 *vs.* 2.7 nats) and the grammar size (mean number of rules with probability > 0.05 was 32 *vs.* 59, resp.). However, the classification test showed that two MLGD runs experienced the over-fitting (drop in AuROC from 1.0 to ca. 0.9). Moreover, with a population of 200 and a longer training, PGE produced comparably sized grammar with a better fit to the training data (2.4 nats) and did not suffer excessive over-specialization.

## 4 DISCUSSION AND CONCLUSIONS

For toy languages, the MLGD algorithm showed improved performance in comparison to PGE in finding grammars that are both concise and well fitting the probability distribution of the training sample. Their discriminative power was also high. It seems plausible that the gain is in the new mutation operator, which facilitates pruning from redundant rules (and variables). A remaining issue with MLGD is a varying quality of solutions from different runs. It can also be noted that in some cases keeping grammar larger than would be necessary for the non-probabilistic case resulted in improved probability distribution. Indeed, while the probability normalization, which makes grammars proper, implies that the decreasing number of rules has some positive effect on the probability distribution, this effect is not unconditional. In fact, a precise modeling of the probability mass distribution may require more non-terminals and rules imposing a trade-off between discriminative power and intelligibility of grammars. This may also hint why improvement achieved with the new scheme was less pronounced on the CaMn sample. These sequences share one length, which is more difficult to model with a probabilistic grammar. While the problem could be addressed with adding more variables and rules (see [11]), this solution may not harmonize well with the mutation operator of MLGD. Other differences between the CaMn and toy samples include a larger alphabet and higher syntax complexity of the former. This almost certainly makes the assumed covering grammar underpowered to fully capture the characteristics of the CaMn sample. It remains to investigate performance of MLGD in such conditions more thoroughly. Intended future work includes also comparing differentiating features of the both learning schemes one-by-one. Moreover, we will consider approaches to augmenting our generic covering grammars with some auxiliary variables and limited number of associated rules, in order to improve modeling probability distribution without impeding the learning process and decreasing readability of the grammars.

## ACKNOWLEDGMENTS

This research has been partially funded by National Science Centre, Poland [2015/17/D/ST6/04054] and supported by Wroclaw Centre for Networking and Supercomputing [98] and E-SCIENCE.PL.

## REFERENCES
[1] K. Lari and S.J. Young. The estimation of stochastic context-free grammars using the inside-outside algorithm. *Computer Speech & Language*, 4(1):35, 1990.
[2] Glenn Carroll and Eugene Charniak. Two experiments on learning probabilistic dependency grammars from corpora. In *The Workshop on Statistically-Based Natural Language Programming Techniques*, pages 1–13. AAAI, 1992.
[3] Bill Keller and Rudi Lutz. Evolutionary induction of stochastic context free grammars. *Pattern Recognition*, 38(9):1393 – 1406, 2005.
[4] Kaan Tariman. Genetic algorithms for stochastic context-free grammar parameter estimation. Master's thesis, The University of Georgia, United States, 2004.
[5] W Dyrka and J-C Nebel. A stochastic context free grammar based framework for analysis of protein sequences. *BMC Bioinformatics*, 10:323, 2009.
[6] W Dyrka, M Pyzik, F Coste, and H Talibart. Estimating probabilistic context-free grammars for proteins using contact map constraints. *PeerJ*, 7:e6559, 2019.
[7] R. Kowalski. Maszynowe uczenie gramatycznych deskryptorów sekwencji białkowych, 2019. Engineer's thesis.
[8] Matthew Wall. Matthew's GAlib: A C++ genetic algorithm library. http://lancet.mit.edu/ga, 2005.
[9] Y Sakakibara and M Kondo. Ga-based learning of context-free grammars using tabular representations. In *Proceedings of 16th ICML*, pages 354–360, 1999.
[10] Mikaël Mayer and Jad Hamza. Optimal test sets for context-free languages. *CoRR*, abs/1611.06703, 2016.
[11] M. T. Johnson. Capacity and complexity of hmm duration modeling techniques. *IEEE Signal Processing Letters*, 12(5):407–410, May 2005.